\documentclass[letterpaper]{article} % DO NOT CHANGE THIS
\usepackage{aaai23}  % DO NOT CHANGE THIS
\usepackage{times}  % DO NOT CHANGE THIS
\usepackage{helvet}  % DO NOT CHANGE THIS
\usepackage{courier}  % DO NOT CHANGE THIS
\usepackage[hyphens]{url}  % DO NOT CHANGE THIS
\usepackage{graphicx} % DO NOT CHANGE THIS
\usepackage{natbib}  % DO NOT CHANGE THIS AND DO NOT ADD ANY OPTIONS TO IT
\usepackage{caption} % DO NOT CHANGE THIS AND DO NOT ADD ANY OPTIONS TO IT
\usepackage{algorithm}
\usepackage{algorithmic}

\usepackage{multirow}
\usepackage{color}
\usepackage[normalem]{ulem}
\useunder{\uline}{\ul}{}
\usepackage{paralist}

\newcommand{\dataname}[1]{\textsc{HateMM}}

\usepackage{newfloat}
\usepackage{listings}
\DeclareCaptionStyle{ruled}{labelfont=normalfont,labelsep=colon,strut=off} % DO NOT CHANGE THIS
\floatstyle{ruled}
\newfloat{listing}{tb}{lst}{}
\floatname{listing}{Listing}
\title{HateMM: A Multi-Modal Dataset for Hate Video Classification}
\author {
    Mithun Das\textsuperscript{\rm1}, 
    Rohit Raj\textsuperscript{\rm1}, 
    Punyajoy Saha\textsuperscript{\rm1}, 
    Binny Mathew\textsuperscript{\rm1}, 
    Manish Gupta\textsuperscript{\rm2}, 
    Animesh Mukherjee \textsuperscript{\rm1}
}
\affiliations {
    \textsuperscript{\rm 1} Indian Institute of Technology (IIT), Kharagpur\\
    \textsuperscript{\rm 2} Microsoft, India\\
    \{mithundas,rrohit2901,punyajoys,binnymathew\}@iitkgp.ac.in,\\ gmanish@microsoft.com, animeshm@cse.iitkgp.ac.in
}

\usepackage{bibentry}
\begin{document}

\maketitle

\begin{abstract}
Hate speech has become one of the most significant issues in modern society, having implications in both the online and the offline world. Due to this, hate speech research has recently gained a lot of traction. However, most of the work has primarily focused on text media with relatively little work on images and even lesser on videos.
Thus, early stage automated video moderation techniques are needed to handle the videos that are being uploaded to keep the platform safe and healthy. With a view to detect and remove hateful content from the video sharing platforms, our work focuses on hate video detection using multi-modalities. To this end, we curate $\sim43$ hours of videos from BitChute and manually annotate them as hate or non-hate, along with the frame spans which could explain the labelling decision. To collect the relevant videos we harnessed search keywords from hate lexicons. We observe various cues in images and audio of hateful videos. Further, we build deep learning multi-modal models to classify the hate videos and observe that using all the modalities of the videos improves the overall hate speech detection performance (accuracy=0.798, macro F1-score=0.790) by $\sim5.7$\% compared to the best uni-modal model in terms of macro F1 score. In summary, our work takes the first step toward understanding and modeling hateful videos on video hosting platforms such as BitChute.

\end{abstract}

\section{Introduction}
\textbf{Disclaimer:} The article contains material that many will find offensive or hateful; however this cannot be avoided owing to the nature of the work. \\
Social media platforms allow users to publish content themselves. With 82\% of consumer Internet traffic expected to be video~\cite{wilson_2022} in 2023, video hosting platforms like YouTube, Dailymotion etc. have emerged as a major source of information. On YouTube itself, people watch more than a billion hours of video every day\footnote{\url{https://www.youtube.com/intl/en-GB/about/press/}}. The viral nature of such videos is a double-edged sword; on one hand it can help very quick news propagation, on the other hand it can spread hate or misinformation quickly as well. These videos cover a wide-range of topics and while most of the content on YouTube is harmless, there are videos which violate the community guidelines~\cite{connor_2021}. This issue is more severe for some of the alternative video hosting platforms like BitChute\footnote{\label{footnoteBitchute}\url{https://www.bitchute.com/}}, Odysee\footnote{\url{https://odysee.com}} etc. While platforms like YouTube, Facebook, Twitter have strong moderation policies in place, these Alt-Tech platforms\footnote{\url{https://en.wikipedia.org/wiki/Alt-tech}} allow users to post any content with little to no moderation. The non-removal of such content could be detrimental for the users and the website as a whole. It could lead to a hostile environment with echo-chambers of hateful users. It could also lead to a loss of revenue as well as attract fines~\cite{troianovski_schechner_2017} and lawsuits.

Some platforms employ several human moderators to find the harmful content and remove them from their site. However, given the amount of content posted daily, it is a very daunting challenge. For example, Facebook employs around 15K moderators to review content flagged by its AI and users~\cite{koetsier_2021} and makes around 300K content moderation mistakes every day. Further, the moderators themselves are at the risk of 
emotional and psychological trauma~\cite{newton_2019}.  This issue is further exacerbated by laws which require the platforms to remove hateful content within a fixed period of time. Failure to abide by these regulations could lead to fines~\cite{troianovski_schechner_2017}. 
While platforms like YouTube have machine learning algorithms in place to detect hateful content, smaller platforms might not have the revenue/technology to develop datasets/models for hate speech detection in videos. Thus, there is a need to develop open efficient models which could detect hate speech in videos. However, the current research on hate speech is mostly focused on text based models~\cite{badjatiya2017deep,cheng2020unsupervised,juuti2020little,kennedy2020contextualizing,parikh2021categorizing,das2022data} with very few image-based ones~\cite{yang2019exploring,das2020detecting,gomez2020exploring,kiela2020hateful}. 
Detecting hateful actions in videos needs leveraging a combination of multi-frame video processing and speech processing signals, and thus image-based hate detection methods cannot be directly adapted.

\noindent\textbf{Research objectives and contributions}: In this paper, we take a step toward an end-to-end solution for this novel problem setting. We release \dataname{}, a collection of videos annotated for hate speech. The dataset contains $\sim43$ hours of videos composed of a total of $\sim144$K frames. We make the \dataname{} dataset public\footnote{\url{https://doi.org/10.5281/zenodo.7799469}} to promote further research in multi-modal hate speech detection.
We rely on BitChute for our data collection as it has low content moderation. Launched in 2017, BitChute serves as a video hosting and sharing platform similar to YouTube and is quite popular among far-right users.

Overall, we make the following \textbf{contributions}.

\begin{compactitem}
    \item We curate one of the largest known datasets of hateful videos consisting of 1083 videos spanning $\sim43$ hours and $\sim144$K frames. Each individual video was annotated as hateful or not, along with the frame spans which justify the labelling decision. The average time taken by the annotators to label a single video was approximately twice the video duration.
    \item We develop detection models using three different modalities (text, audio and video) individually as well as jointly\footnote{The source code of the baseline models is available at ~\url{https://github.com/hate-alert/HateMM}}. Our best fusion model (BERT $\odot$ ViT $\odot$ MFCC) which combines all the modalities attains a macro F1-Score of 0.790. The precision and the recall for the hate class are 0.742 and 0.758 respectively. Among the individual modalities, transformer encodings of text and video-based features seem to be more effective for detecting hateful videos.

    \item We further perform some preliminary analysis of the importance of each of the modalities. We observe that the text based model is successful when the transcript is relatively clean. The audio based model is most effective when there is shouting and expression of aggression in the video. Finally, the vision based model works the best if there is evidence of visual hateful activity with presence of victim in the video.
    \item As a last step, we analyze the performance based on the frame spans and observe that the text-based and vision-based models can leverage this information the best. Besides, the vision-based model performs the best in case the hate target in the video are `Blacks' or `Jews', while the text-based model does very well on the `Other' target communities.  
\end{compactitem}

%The remainder of the paper is organized as follows. Immediately after this, we cover some of the related research. Post this, we discuss the annotation guidelines for the creation of \dataname{} dataset and its basic statistics. Next, we present the different models used for the hate speech video classification task. Further, we present the experimental setup and the results. Finally, we conclude our paper with a brief summary.

\begin{figure}[h]
  \centering
  \includegraphics[width=0.6\linewidth]{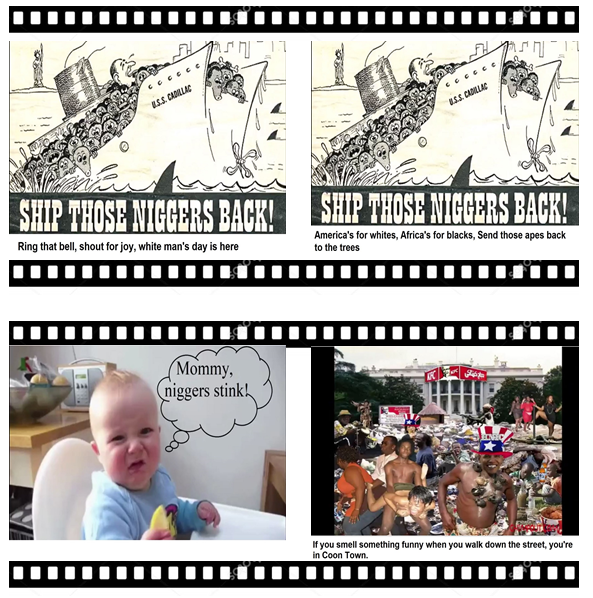}
  \caption{Examples of hate videos.}
  \label{fig:HateVideo}
\end{figure}

\section{Related Work}
\label{sec:related_works}

With the huge availability of multi-modal data, multi-modal deep learning has been harnessed to improve the accuracy for various tasks like visual question answering~\cite{Singh_2019_CVPR}, fake news/rumour detection~\cite{khattar2019mvae}, etc. 
Recently, multi-modal hate speech detection has become popular where text posts are combined with extra contexts like user and network information~\cite{cheng2020unsupervised,founta2019unified} or images~\cite{yang2019exploring,das2020detecting,gomez2020exploring,kiela2020hateful} to improve detection accuracy. Such multi-modal schemes typically use unimodal methods like CNNs, LSTMs or BERT to encode text and deep CNNs like ResNet or InceptionV3 to encode images, and then perform multi-modal fusion using simple concatenation, gated summation, bilinear transformation, or attention-based methods. Multi-modal bitransformers like ViLBERT and Visual BERT have also been applied~\cite{kiela2020hateful}. 

There is almost no work on the detection of offensive/hate videos barring the following three -- for Portuguese~\cite{alcantara2020offensive} and English~\cite{wu2020detection,rana2022emotion}. Nevertheless, the first two works~\cite{alcantara2020offensive,wu2020detection} only consider textual features for their classification purpose by extracting the transcript. Further, the size of the annotated dataset is less than 500 videos. The work done by Rana et al.~\shortcite{rana2022emotion} considered both textual and audio features, though the dataset is not publicly available, the data curation and annotation steps are not fully described and the dataset statistics are not precisely revealed. Unlike our dataset, they choose videos where the speech is clear; in contrast, we did not have any such constraint since hateful content can be as well expressed in only visual form without having any associated speech. To the best of our knowledge, we are the first to experiment with multi-modal hate video detection, where we leverage all three data modalities -- text, audio, and video. Our annotation is far richer and larger compared to the state-of-the-art in order to appropriately leverage all the modes. We believe that our dataset and the benchmark models trained on it will help the moderators identify genuine hateful cases while reducing false alarms.

\section{\dataname{} Dataset}
\label{sec:dataset}

\subsection{The BitChute Platform}
BitChute is a social video-hosting platform with low content moderation launched as an alternative to YouTube~\cite{trujillo2020bitchute}. The website launched in 2017 is gaining popularity and is becoming a ``haven'' for far-right users. BitChute has high prevalence of hateful content and hosts several content producers who were banned from traditional and moderated platforms~\cite{labarbera_2020}.
%\bm{We can maybe talk about some of the results from previous research or policies at bitchute}\am{Add some references to past works on this platform.}

\subsection{Data Collection}
% Starting in 2017, Bitchute serves as an video hosting and sharing platform similar to Youtube. We rely on Bitchute for our data collection as it has low content moderation. 
To sample the videos for annotation we used lexicons from~\cite{mathew2020hate} that studied Gab and other alt-right platforms. These lexicons consist of derogatory keywords/slurs targeting different protected communities.

Each of the keywords is used to search on BitChute; the links returned are added to a database. In total, we collected $\sim8K$ links. Next we download the videos using 
BitChute-dl software\footnote{\url{https://pypi.org/project/bitchute-dl/}}. While downloading we did not find the videos for 25\% of the links. Further few videos were corrupted as well. Finally we end up with $\sim6K$ videos.

\subsection{Annotation Guidelines}

The labeling scheme stated below constitute the main guidelines for the annotators, while a codebook ensured common understanding of the label descriptions. We construct our codebook (which consists the annotation guidelines) for identifying hateful content on the YouTube policy of hate speech\footnote{\url{https://support.google.com/youtube/answer/2801939?hl=en}}. We consider a video as hateful if --

\begin{quote}
\textit{``It promotes discrimination or disparages or humiliates an individual or group of people on the basis of the race, ethnicity, or ethnic origin, nationality, religion, disability, age, veteran status, sexual orientation, gender identity etc.''}
\end{quote}

In addition, we also ask the annotators to mark the parts (i.e., frame spans) of a hate video which they felt are hateful (as rationales) and the communities the video targets. We believe that the rationales can later serve as an explainability signal and targets can be used to measure if the detection algorithms are getting biased toward some targets in the lines of what has been presented in~\cite{mathew2020hatexplain}.

\subsection{Annotation Process}

\subsubsection{Training the annotators} The annotation process was led by two PhD students as expert annotators and performed by four under-graduate students who were novice annotators. All the undergraduate students are computer science majors, All of them participated voluntarily for the task with complete consent and were rewarded through an online gift card at the end of the task. Both the expert annotators had experience in working with harmful content in social media. In order to train the annotators we needed a pilot gold tagged dataset. To this end, the expert annotators initially annotated 30 videos. The initial set consisted of 20 hate videos and 10 non-hate videos. We gave these 30 videos to the undergraduate annotators who annotated these based on the annotation codebook. Once they finished their annotations we discussed the incorrect cases with them to improve their annotation skills.

\subsubsection{Annotations in batch mode} Subsequent to the above training, we released a set of 30 videos per week in a batch mode. Being aware that while annotating hate videos, annotators can have ``negative psychological effects''~\cite{ybarra2006examining}, we advised them to take at least 10 minutes break after the annotation of each video. We further imposed an additional constraint that no more than 10 videos should be annotated per day. Finally, we also had regular meetings with them to ensure the annotations did not have any adverse effect on their mental health.

\subsubsection{Annotation tool} %One challenge of the annotation process was to have a proper annotation tool. 
\begin{figure}[h]
  \centering
  \includegraphics[width=0.9\linewidth]{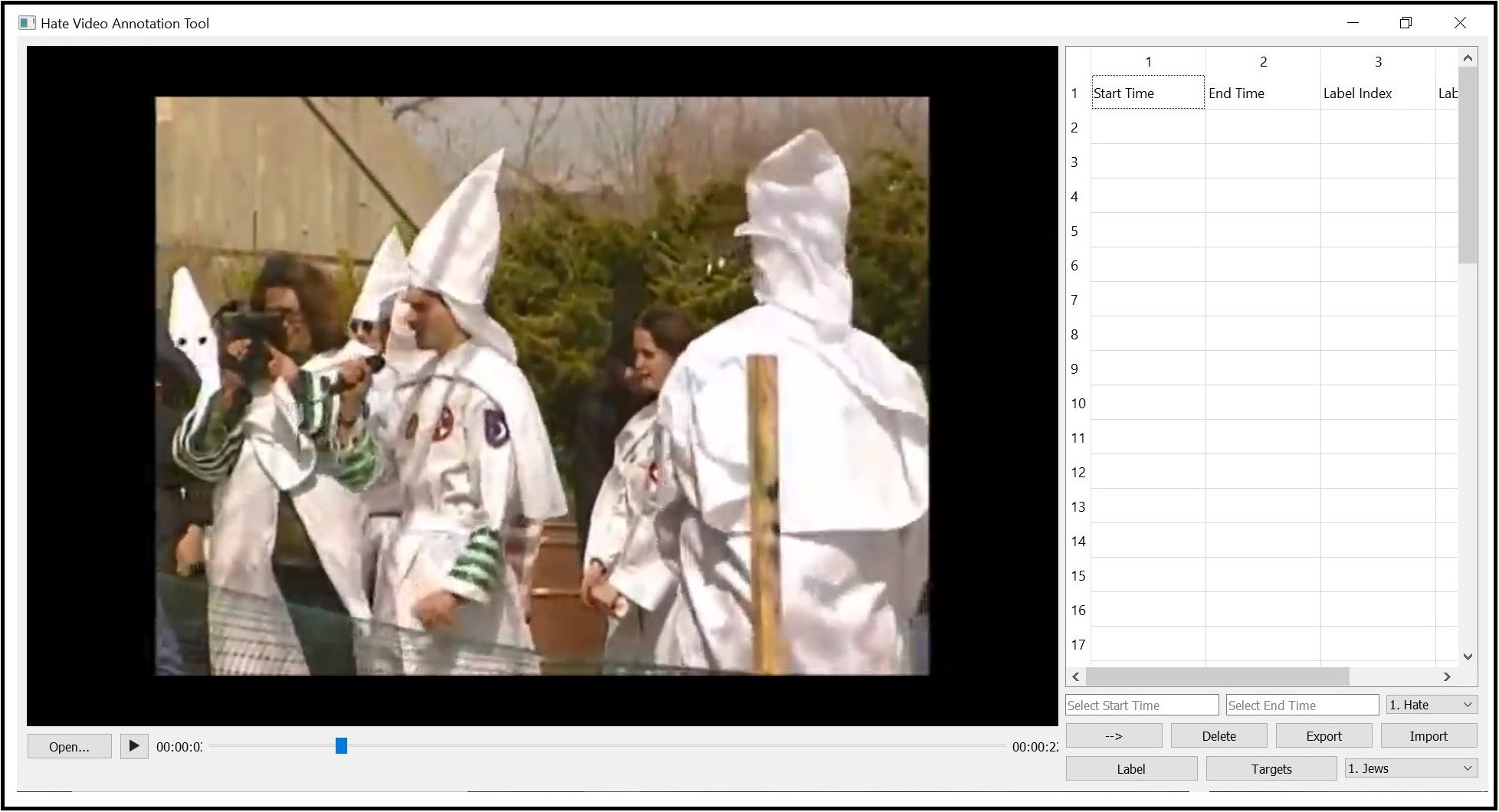}
  \caption{Snapshot of the (hate) video annotation tool.}
  \label{fig:Annotation_Tools}
\end{figure}

%As an initial choice for the annotation tool, we resorted to 
Off-the-shelf tools like Toloka\footnote{\url{https://toloka.ai/ml/computer-vision}} and ANVIL ~\cite{ANVIL} allow annotations for tasks like object detection, but they do not support annotations of any kind of spans in videos. 
PAVS\footnote{\url{https://github.com/kevalvc/Python-Annotator-for-VideoS}} only allows span selection but fortunately it is open-source. %the code was open-sourced. 
Consequently, we modified PAVS to support span annotations, hate or not video annotation, and target community labeling. We shall make our annotation tool (see a snapshot of the tool in Figure~\ref{fig:Annotation_Tools}) public to facilitate further research.

Each video was annotated by two independent annotators. They were instructed to watch the complete video and based on the guidelines provided, select the appropriate class (hate or non hate). Average time required by the annotators to annotate a video was approximately \textit{twice} the video duration. The Cohen's kappa for the inter-annotator agreement was $\kappa$=0.625. On completion of each batch of annotation, if there was a mismatch between the two annotators, one of the expert annotators annotated the same video to break the tie. This yielded a final dataset of 431 hate and 652 non-hate videos and constitutes our set of a total of 1083 labelled instances.

%\mg{In Figure~\ref{fig:HateVideo}, can we try to have some variety? Both videos seem to highlight ``niggers''.} \md{In this set the hateful videos we have are majorly targetting blacks.}

%\punyajoy{Add about the rounds and challenges}

\begin{table}
\centering
\small
\begin{tabular}{l|c|c|c|}
\cline{2-4}
& \textbf{Hate} & \textbf{Non Hate} & \textbf{Total} \\ \hline
\multicolumn{1}{|l|}{\textbf{Count}}                    & 431 (39.8\%)   & 652 (60.2\%)       & 1083           \\ \hline
\multicolumn{1}{|l|}{\textbf{Total len (hrs)}}        & 18.39         & 24.87             & 43.26          \\ \hline
\multicolumn{1}{|l|}{\textbf{$\mu$ video len}}       & 2.56 ± 1.69   & 2.28 ± 4.77       & 2.40 ± 3.86    \\ \hline
\multicolumn{1}{|l|}{\textbf{$\mu$  rationale len}}     & 1.71 ± 1.27    & -                 & -              \\ \hline
%\multicolumn{1}{|l|}{\textbf{Total frames}}             & 66.6K         & 90.1K             & 156.7K         \\ \hline
\multicolumn{1}{|l|}{\textbf{$\mu$  \#frames}}              & 154           & 137               & 144            \\ \hline
\multicolumn{1}{|l|}{\textbf{$\mu$  \#words}} & 228           & 209              & 217            \\ \hline
\end{tabular}
\caption{Basic statistics of the \dataname{} dataset. Frames were sampled per second. Video and rationale length are in minutes. len: length. hrs: hours. $\mu$ : Mean.}
\label{tab:basic-stat}
\end{table}

\subsection{Dataset Statistics}
Our final dataset contains 1083 videos spanning over $\sim43$ hours of content. On average the videos are $\sim2.40$ mins in length with hate videos being slightly longer at an average of $\sim2.56$ mins. Overall we are able to curate a roughly balanced dataset with 39.8\% of the samples labelled as hate. The class balance is better than many of the textual hate speech datasets. For each video, we also get the audio transcribed using Vosk offline speech recognition\footnote{\url{https://alphacephei.com/vosk/}} tool. There are $\sim217$ words in the transcripts on average. Further, the number of words in the transcripts of hate videos are slightly higher ($\sim228$) compared to transcripts of non-hate videos. The other important statistics of the dataset are noted in Table~\ref{tab:basic-stat}. Snapshots of some examples hate videos are shown in Figure~\ref{fig:HateVideo}.

\begin{figure}[h]
    \centering
    \includegraphics[width=0.8\linewidth]{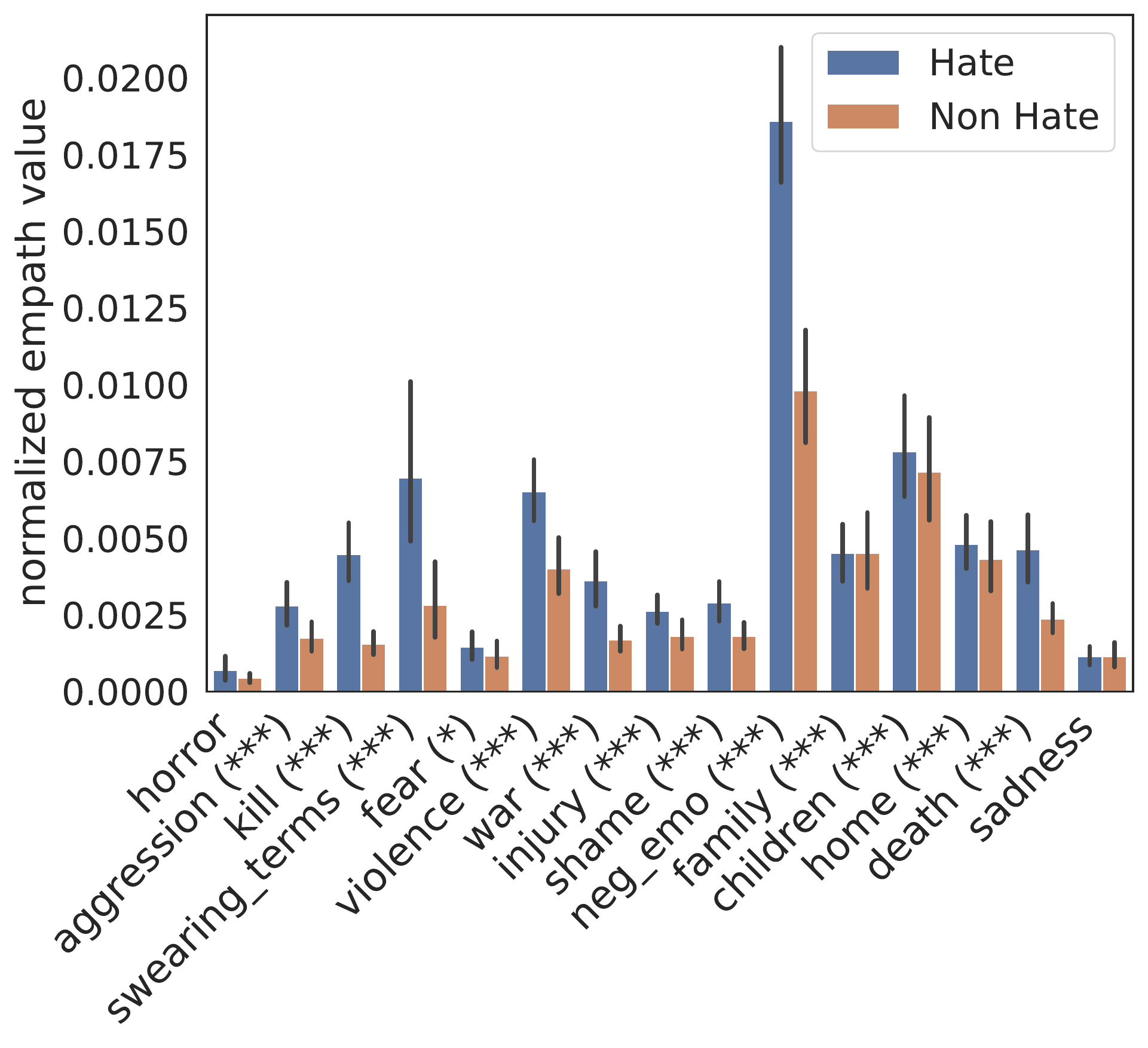}
    \caption{Lexical analysis of video transcripts using Empath. We report the mean values for several categories of Empath. Hate video transcripts scored significantly high in categories like `aggression', `swearing terms', `violence' and `negative emotion'. For each category, we use the Mann-Whitney U test and show the significance levels ***$(p < 0.0001)$, **$(p < 0.001)$, *$(p < 0.01)$. }
    \label{fig:empath}
\end{figure}

\subsection{Dataset Analysis}

\subsubsection{Empath analysis}
In order to understand the dataset better, we identify important lexical categories present in the video transcripts using Empath~\cite{fast2016empath}, which has 189 such pre-built categories.  First, we select 70 categories ignoring the irrelevant topics to hate speech, e.g., technology and entertainment. We report the top 15 significantly different categories in Figure~\ref{fig:empath}. Hate video transcripts scored significantly high in categories like `aggression', `swearing terms', `violence' and `negative emotion'.

\begin{figure}[h]
    \centering
    \includegraphics[width=0.48\linewidth]{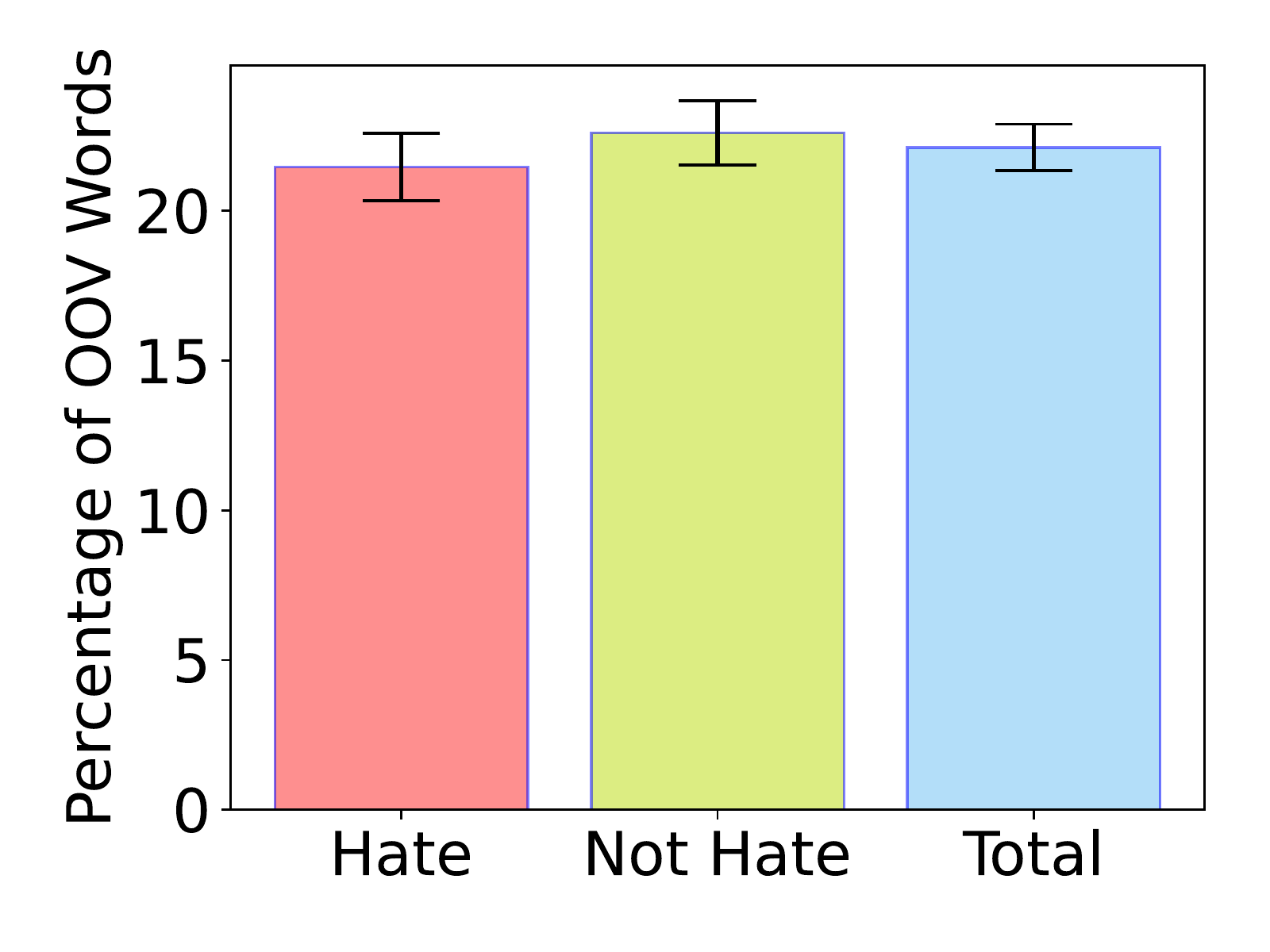}
    \caption{Percentage of OOV words present in the dataset.}
    \label{fig:oov_word}
\end{figure}

\subsubsection{OOV words} 
To understand the extent of noise in the dataset (and the transcript quality), we calculate the percentage of Out-of-Vocabulary (OOV) words present in both the `hate' and the `non-hate' classes. For this purpose, we use the PyEnchant dictionary\footnote{\url{https://github.com/pyenchant/pyenchant}}, which is Python's spellchecking dictionary, to identify the words that are not present in the standard English library. Figure \ref{fig:oov_word} shows the \% OOV words per video (transcript) for both the `hate' and the `non-hate' classes. The plot shows that for both the `hate' and the `non-hate' classes, the mean percentage of OOV words is almost 22\%.

\begin{figure}[h]
    \centering
    \includegraphics[width=0.4\linewidth]{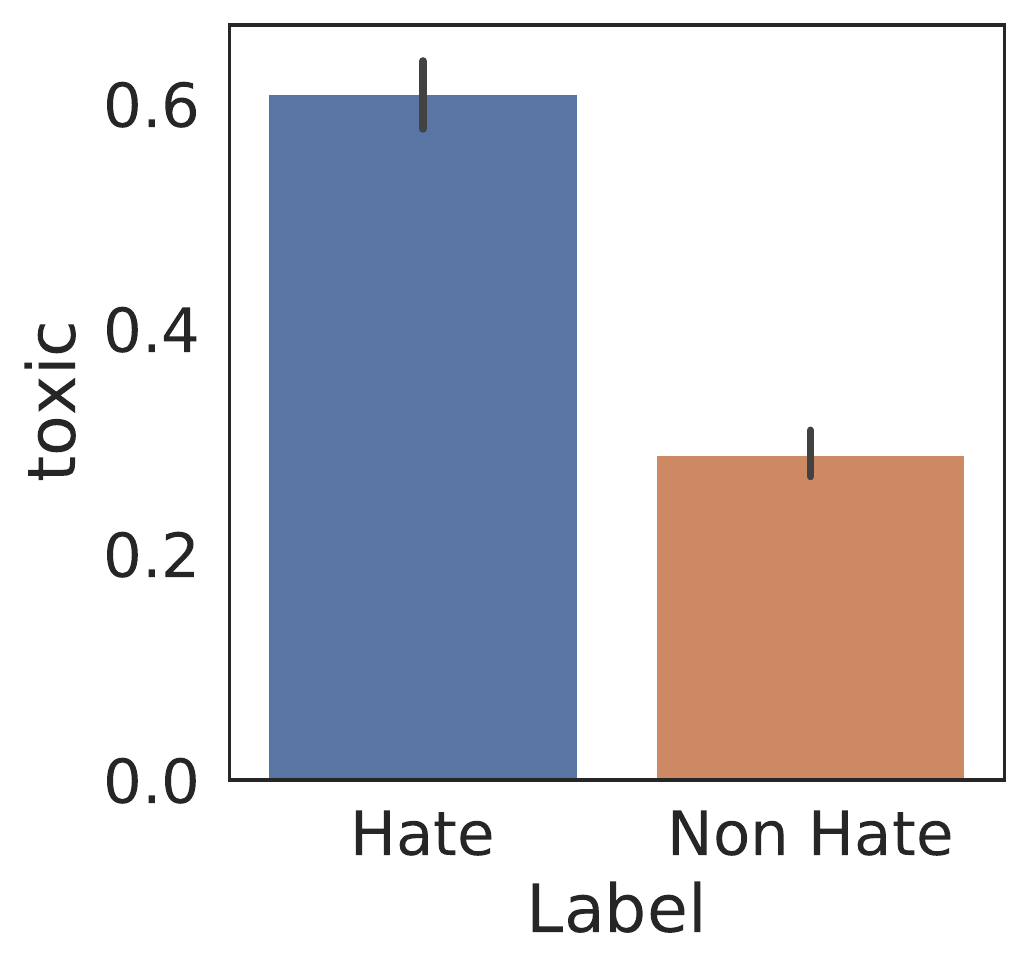}
    \caption{Toxicity comparison based on Perspective API. The results are significant at $p<0.0001$ based on Mann-Whitney U test.}
    \label{fig:toxicity}
\end{figure}

\subsubsection{Toxicity score} 
One easy solution to detect toxic videos could be to use Google's Perspective API\footnote{\url{https://www.perspectiveapi.com/}}
%~\cite{perspective} 
on the transcript; hence we measured the toxicity of the transcripts. As shown in Figure~\ref{fig:toxicity}, 
%the average toxicity score for the hate video transcripts is $\sim0.61$. While comparing the categories, we find that 
hate video transcripts have 
%\mg{should we say slightly higher -- it is actually twice, right?}\punyajoy{corrected} 
almost twice the toxicity ($\sim0.61$) compared to the non-hate video transcripts ($\sim0.28$).
However, relying on transcripts has its own drawbacks. As discussed above, the transcripts, in general, are quite noisy, and this indicates why transcripts alone might not be sufficient for hate video classification.
%words are almost 22\% in the transcript and negative sentiment videos are more in hateful data points. More detail can be found in the appendix.

\begin{figure}[h]
    \centering
    \includegraphics[width=0.6\linewidth]{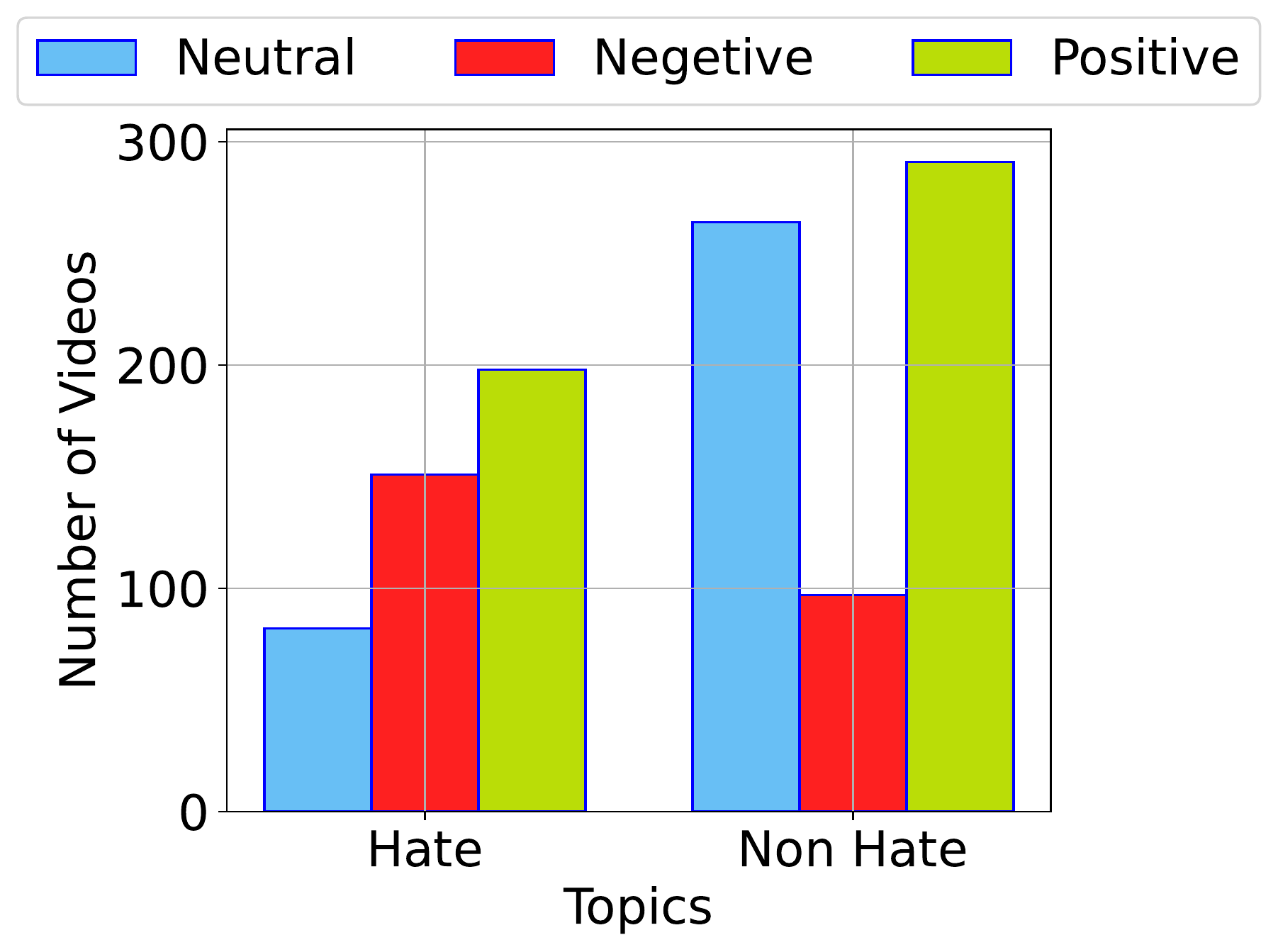}
    \caption{Number of posts having different sentiments in hate and non-hate category.}
    \label{fig:video_topic_sentiment}
\end{figure}

\subsubsection{Sentiments} We also measure the sentiment associated with the videos that we have annotated. Sentiment analysis is used to identify the associated feelings/emotions within a text. Sentiment analysis includes three types of polarity: negative, neutral, and positive. In this study, the word-based method was used and the polarity of each transcript was determined by the score from -1 to 1 according to the word used. A negative score means a negative sentiment, and a positive score means a positive sentiment. Sentiment analysis was carried out using TextBlob API\footnote{\url{https://textblob.readthedocs.io/en/dev/}}. Figure ~\ref{fig:video_topic_sentiment} represents the number of videos having neutral, negative and positive sentiment for both video categories. We observe that though overall videos with positive sentiment are more, `hate' video transcripts have more negative sentiment compared to `non-hate' videos.

\begin{table}
\centering
\begin{tabular}{|c|c|}
\hline
\textbf{Hate} & \textbf{Non Hate} \\ \hline
k k k k       & joe               \\
k k k         & joe rogan         \\
k k           & george zimmerman  \\
uncle sam     & joe biden         \\
jack          & bush              \\
joe           & chris             \\
klan          & mug               \\
robert hours  & mike              \\
max           & eric              \\
k k k k k     & johnson           \\ \hline
\end{tabular}
\caption{Most frequent PERSON entities in Hate and Non Hate classes}
\label{tab:nerPerson}
\end{table}

\begin{figure}[h]
    \centering
    \includegraphics[width=\linewidth]{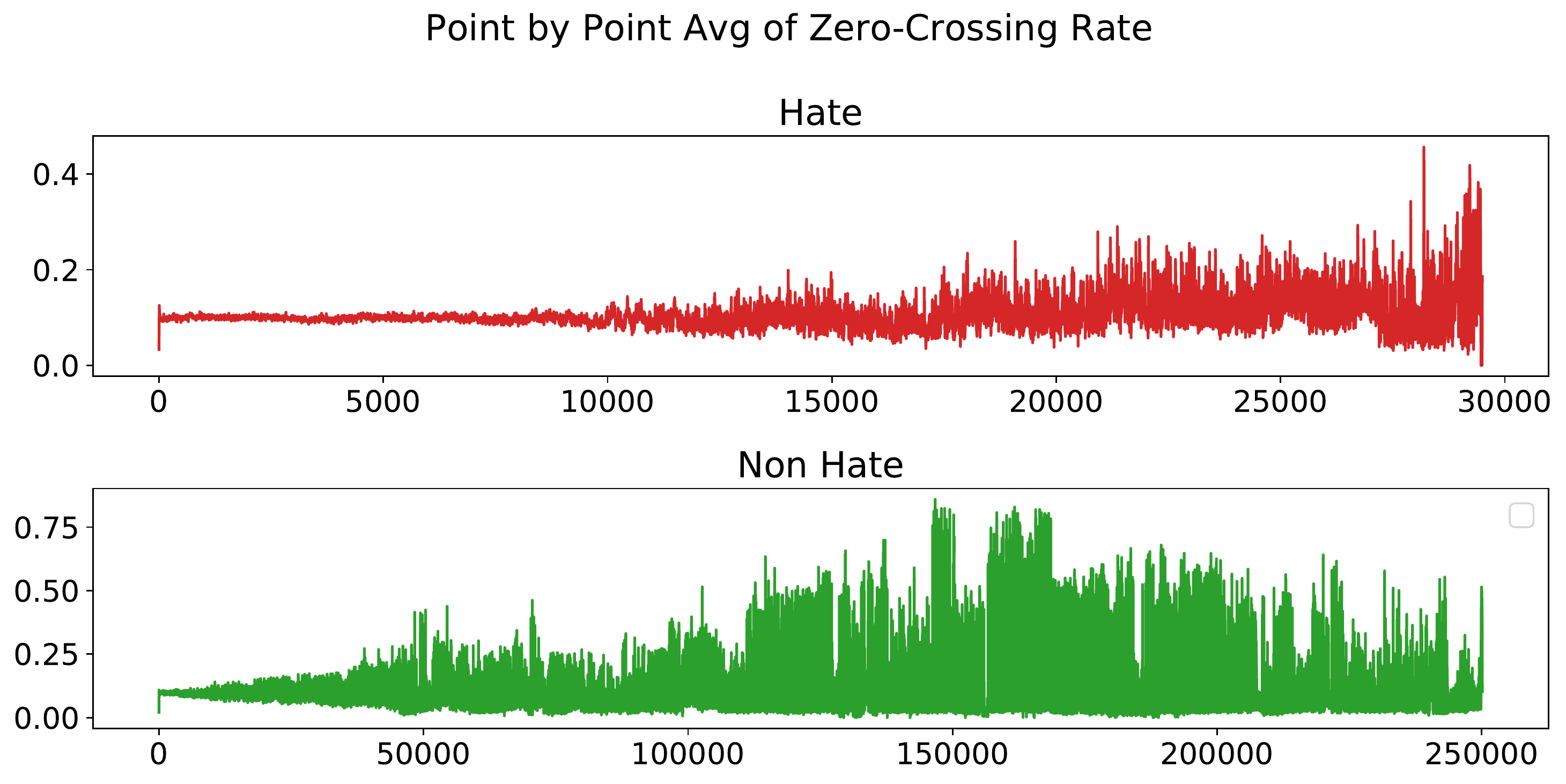}
    \caption{Zero Crossing rate for the hate and non-hate videos.}
    \label{fig:zero_figure}
\end{figure}

\subsubsection{NER analysis} 
We also analyzed the named entities associated with the transcript to 
find out if the distribution of different NER tags is different for hate and non hate classes. To this purpose, we use the spacy library~\footnote{\url{https://spacy.io/}}, which provides a set of entity tags. We observed that for the PERSON tag, the normalized number of named entities are more in the hate class than the non hate class. We further inspected the most frequent entities associated with the entity type PERSON which are noted in Table~\ref{tab:nerPerson}. For the hate class, phrases like `k k k' are very common; this possibly corresponds to the KKK\footnote{\label{footnoteKKK}\url{https://en.wikipedia.org/wiki/Ku_Klux_Klan}}, an American white supremacist terrorist and hate group whose primary targets are African Americans, Jews, etc.

\subsubsection{Audio analysis} 
We also analyze the audio signal associated with each video. Specifically, we calculate the Zero-Crossing Rate (ZCR), Spectral Bandwidth, Root Mean Square (RMS) Energy and report the mean for all the audios. 

In Figure~\ref{fig:zero_figure} we plot the time series of the ZCR averaged over all audio files in the two respective classes. We observe that the plots are distinctly different for the hate and the non hate classes. It is well known that ZCR can be interpreted as a measure of the noisiness of a signal, and higher values indicate more noisiness of the audio signal. This indicates that while the noise level is roughly uniformly spread over the whole time series for the non hate videos, for the hate videos this is predominantly flat and only appear to go up toward the end of the time series indicating that the hate videos are possibly crafted to have better quality audio signal. The same results are also observed for the Spectral Bandwidth (data not produced for brevity).

\begin{figure}[h]
    \centering
    \includegraphics[width=\linewidth]{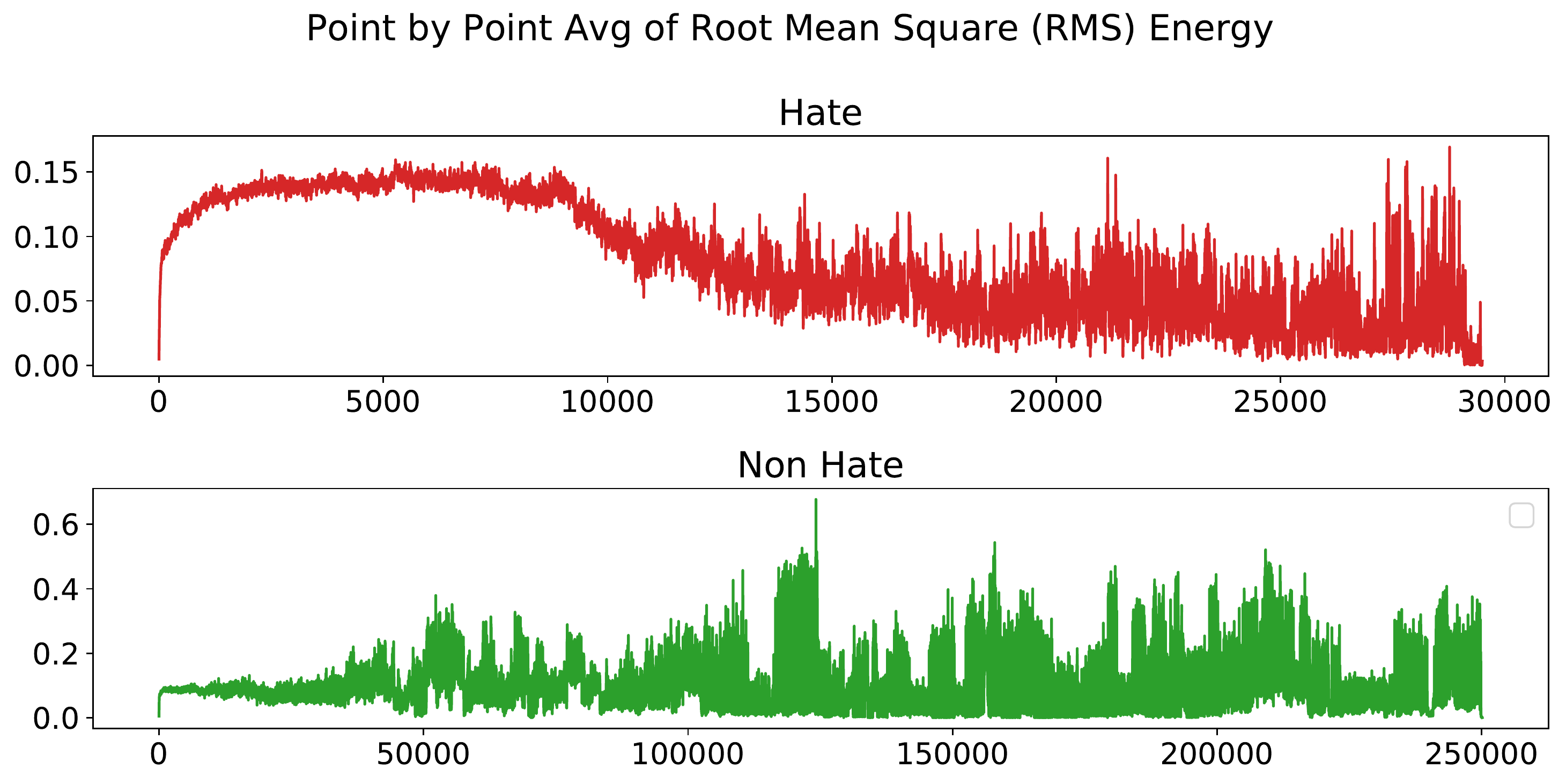}
    \caption{Root Mean Square (RMS) Energy of Hate and non-hate videos.}
    \label{fig:rms_figure}
\end{figure}

In Figure~\ref{fig:rms_figure} we present the RMS Energy plot for the hate and the non hate videos averaged over all the audio files in the respective classes. RMS energy is helpful in estimating the average loudness of an audio track. We observe that hate videos are louder only in the initial part of the time series unlike for non hate videos. A manual inspection showed that in many hate video there are instances of shouting which possibly manifests in the form of high loudness.

\subsubsection{Video analysis} 

We further attempt to analyze what kind of objects are mainly present in the videos. To this purpose, we use the ImageAI~\footnote{\url{https://imageai.readthedocs.io/}} object detection package. For each hate and non-hate video, we randomly select 20 frames and extract all the objects associated with the frames. We assume if an object has been seen in any frame of a given video, it would mean that the object is present in that video. We observe that 45\% of the time, the object ``person" appears in hateful videos, whereas 59\% of the time, the object ``person" appears in non hateful videos. In 
the hateful videos we observe use of important religious persons like a Jewish rabbi at the background with a lot of hateful text embedded on them. Similarly the black persons detected are often associated with dirtiness and food mongering.  Further, for the hate videos, we observe `stop sign' and certain play items like `teddy bear', `kite' and `sports ball'. A manual inspection shows that these play items are mostly `cartoon-ish' figures used to mock a target community.

\noindent Overall, in this section we observe that all three modalities -- text, audio and video have certain latent indicators that should be helpful in differentiating the hate from the non-hate class of videos. This observation, as we shall see, is corroborated by the superior performance of the joint model as observed in section~\ref{sec:results}.

\section{Methodology}
\label{sec:methods}

This section discusses the pre-processing steps and models we implemented for hate video detection.

\subsection{Problem Formulation}

We formulate the hate video detection problem in this paper as follows. Given a video $\textbf{\textit{V}}$, the task can be represented as a binary classification problem. Each video is to be classified as hate ($y = 1$) or non-hate ($y = 0$). A video $\textbf{\textit{V}}$ can be expressed as a sequence of frames, i.e., $\textbf{\textit{F}} = \{f1, f2, ..,fn\}$, the associated audio $\textbf{\textit{A}}$ and the extracted video transcript $\textbf{\textit{T}} = \{w_1, w_2, ..., w_m\}$, consisting of a sequence of words. We aim to learn such a hate video classifier  $Z:Z(\textbf{\textit{F}}; \textbf{\textit{A}}; \textbf{\textit{T}}) \rightarrow y$, where $y \in \{0,1\}$ is the ground-truth label of a video.

\subsection{Pre-processing}

We remove numbers and special characters from the transcripts and perform text normalization wherever required. For vision-based models, we first sample the video at one frame-per-second and sample 100 such frames for each video. For videos with less than 100 frames, we add an image with white background as a padding. For the videos having more than 100 frames, we uniformly sample 100 frames from the total number of available frames.

\subsection{Text-Based Models}
\noindent\textbf{fastText:} We obtain 300 dimensional fastText~\cite{grave2018learning} embedding of all the video transcripts, pass it through two dense layers of 128 nodes, and finally pass it to the output node for the final prediction. We name this model as \textbf{T1}. \\
\noindent\textbf{LASER:} We obtain 1024 dimensional LASER~\cite{artetxe2019massively} embedding of all the video transcripts, pass it through two dense layers of 128 nodes, and finally provide it to the output node for the final prediction. We name this model as \textbf{T2}. \\
\noindent\textbf{BERT:} We use BERT~\cite{devlin2018bert} since it is known to be highly effective for many text classification tasks, including text-based hate speech detection. For each transcript, we get the CLS embedding and pass it through two dense layers of 128 nodes, and finally provide it to the output node for the final prediction. We call this model as \textbf{T3}. \\
\noindent\textbf{HateXPlain:} We also experiment with another BERT model~\cite{mathew2020hatexplain}. The model is already finetuned on pre-trained BERT using English hate speech data. Since this model has been already finetuned on hate speech data, we expect that it should yield better performance. We denote this model as \textbf{T4}.

\subsection{Audio-Based Models}
 \noindent\textbf{MFCC:}  One of the popular methods for representing audio is the Mel Frequency Cepstral Coefficient (MFCC)~\cite{xu2004hmm} which has been found to be effective for complex tasks like lung sound classification~\cite{jung2021efficiently} and speaker identification~\cite{kalia2020comparative}. We obtain a representation of the audio of our dataset using the MFCC features. To generate the MFCC features, we use the open-source package - Librosa\footnote{\url{https://librosa.org/doc/latest/index.html}} and construct a 40 dimensional vector to represent the audio. These vectors are passed through three fully connected layers to generate the final label. We refer to this model as \textbf{A1}. \\
 \noindent\textbf{AudioVGG19:} We also use waveforms of audios (extracted from the videos) and generate 1000 dimensional feature vectors by using a pre-trained VGG-19 model~\cite{Simonyan15,Grinstein2018AudioST}. Similar to \textbf{A1}, these vectors are passed through three fully connected layers to generate the final label. We call this model \textbf{A2}.

\subsection{Vision-Based Models}

In order to handle the spatial and temporal information in the videos, we consider several vision-based classification models such as 3D-CNN, InceptionV3, Vision Transformer, etc. \\
\noindent\textbf{3D-CNN:} The 3D-CNN~\cite{6165309} model contains two Conv3D and BatchNorm3D layers. After these layers we add ReLU, dropout and maxpool layers to generate the final representation. This is further passed through three fully connected layers to generate the final label (please see experimental setup for further details on the layer sizes). We name this model as \textbf{V1}. \\
 \noindent\textbf{InceptionV3:} We also construct feature vectors by using pre-trained InceptionV3 model~\cite{DBLP:journals/corr/SzegedyVISW15}. We extract a 1000 dimensional feature vector for all the 100 frames and then pass it through an LSTM~\cite{lstm} network, which is finally fed to the output node for classification. We use LSTM to capture the sequential nature of the video frames. We name this model as \textbf{V2}. \\
 \noindent\textbf{Vision Transformer:} In this approach, the image is divided into a sequence of patches and then fed to the a transformer model. Like BERT, the extra learnable [class] token is also prepended with the sequence of patches for the classification task. As our focus is to detect hateful videos, so we cannot use Vision Transformer directly. Like the \textbf{InceptionV3} model, we take 100 frames for each video and pass it through the pre-trained Vision Transformer(ViT)~\cite{dosovitskiy2020image} model to get a 768 dimensional feature vector for each frame and finally pass it through the LSTM network to obtain the prediction. We refer to this model as \textbf{V3}.

\begin{figure}
  \centering
  \includegraphics[width=\linewidth]{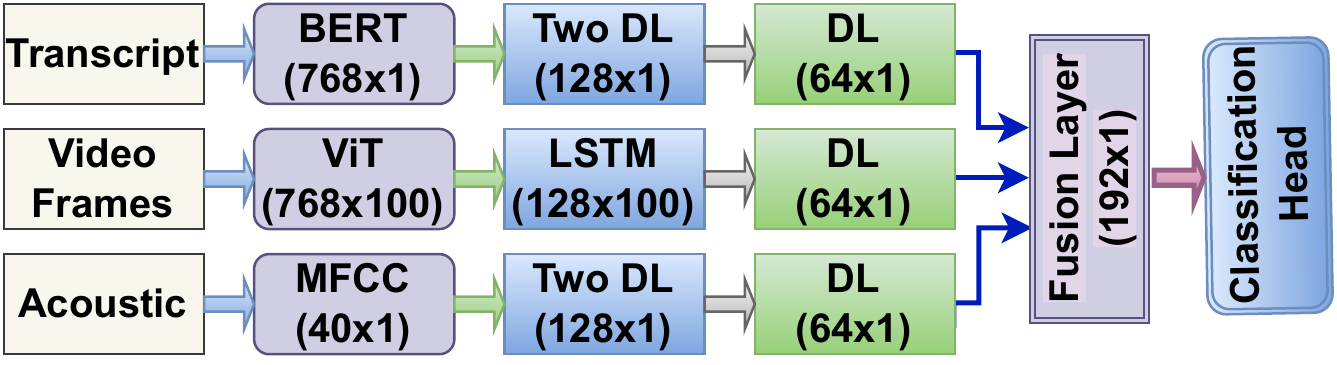}
  \caption{A schematic of the multi-modal model. DL: Dense Layer.}
  \label{fig:fusionModel}
\end{figure}

\subsection{Multi-Modal Hate Video Detection}
The models discussed in the previous subsections are incapable of leveraging the relationship among the features extracted through different modalities (i.e., video, text transcript, and audio). To capture the benefits of all the modalities, we attempt to meaningfully combine the text, audio, and vision-based models. %In this scenario, we fuse the text-based transformer models, vision-based transformer models, and audio-based models discussed above. 
In particular we build the following models -- \textbf{M1} (BERT $\odot$ ViT $\odot$ MFCC), \textbf{M2} (BERT $\odot$ ViT $\odot$ AudioVGG19),  \textbf{M3} (HateXPlain $\odot$  ViT $\odot$ MFCC) and   \textbf{M4} (HateXPlain $\odot$ ViT $\odot$ AudioVGG19). $\odot$ refers to the combination operation of the three modalities through a trainable neural network (aka fusion layer). Figure~\ref{fig:fusionModel} illustrates the overall modeling pipeline.

%------------New Exp ----------------

% \usepackage[normalem]{ulem}
\begin{table*}
\small
\centering
\begin{tabular}{|l|l|l|l|l|l|l|l|c|c|}
\hline
\textbf{Model} & \textbf{Architecture}                    & \textbf{Acc}   & \textbf{M-F1}  & \textbf{F1 (H)} & \textbf{P (H)} & \textbf{R (H)} & & \textbf{\begin{tabular}[c]{@{}c@{}}VL \textless{}= 105 sec\\  (542 videos)\end{tabular}} & \textbf{\begin{tabular}[c]{@{}c@{}}VL \textgreater 105 sec \\ (541 videos)\end{tabular}} \\ \cline{1-7} \cline{9-10} 
\textbf{T1}    & \textbf{fastText} & 0.687    & 0.673    & 0.609     & 0.611    & 0.614    &  & 0.609      & 0.700    \\ \cline{1-7} \cline{9-10} 
\textbf{T2}    & \textbf{LASER}    & 0.730    & 0.720    & 0.668     & 0.655    & 0.686    &  & 0.675      & 0.655    \\ \cline{1-7} \cline{9-10} 
\textbf{T3}    & \textbf{BERT}     & 0.735    & 0.722    & 0.664     & 0.675    & 0.667    &  & 0.672      & 0.708    \\ \cline{1-7} \cline{9-10} 
\textbf{T4}    & \textbf{HXP}      & 0.757    & 0.733    & 0.653     & \textbf{0.753} & 0.577    &  & 0.698      & 0.727    \\ \cline{1-7} \cline{9-10} 
\textbf{A1}    & \textbf{MFCC}     & 0.675    & 0.665    & 0.622     & 0.593    & 0.679    &  & 0.603      & 0.687    \\ \cline{1-7} \cline{9-10} 
\textbf{A2}    & \textbf{AVGG19}   & 0.690    & 0.669    & 0.589     & 0.629    & 0.559    &  & 0.583      & 0.669    \\ \cline{1-7} \cline{9-10} 
\textbf{V1}    & \textbf{3D-CNN}   & 0.674    & 0.653    & 0.571     & 0.619    & 0.547    &  & 0.637      & 0.587    \\ \cline{1-7} \cline{9-10} 
\textbf{V2}    & \textbf{InceptionV3}                     & 0.720    & 0.706    & 0.643     & 0.653    & 0.637    &  & 0.672      & 0.707    \\ \cline{1-7} \cline{9-10} 
\textbf{V3}    & \textbf{ViT}      & 0.748    & 0.733    & 0.672     & 0.695    & 0.656    &  & 0.718      & 0.703    \\ \cline{1-7} \cline{9-10} 
\textbf{M1}    & \textbf{BERT $\odot$ ViT $\odot$ MFCC}   & \textbf{0.798} & \textbf{0.790} & \textbf{0.749}  & {\ul 0.742}    & \textbf{0.758} &  & \textbf{0.772}   & \textbf{0.759} \\ \cline{1-7} \cline{9-10} 
\textbf{M2}    & \textbf{BERT $\odot$ ViT $\odot$ AVGG19} & 0.755    & 0.765    & 0.718     & 0.723    & 0.719    &  & 0.743      & 0.733    \\ \cline{1-7} \cline{9-10} 
\textbf{M3}    & \textbf{HXP $\odot$ ViT $\odot$ MFCC}    & {\ul 0.777}    & {\ul 0.767}    & {\ul 0.720}     & 0.718    & {\ul 0.726}    &  & \underline{0.744}& \underline{0.741}                     \\ \cline{1-7} \cline{9-10} 
\textbf{M4}    & \textbf{HXP$\odot$ ViT $\odot$ AVGG19}   & 0.767    & 0.756    & 0.707     & 0.714    & 0.712    &  & 0.733      & 0.731    \\ \hline
\end{tabular}
\caption{[Left side] Model performance on the task of classification of hate videos. [Right Side] Macro F1 Score with respect to video length (VL) in secs. H: hate class, Acc: accuracy, M-F1: macro-F1, P: precision, R: recall, HXP: HateXplain.}
\label{tab:performanceMetricNew}
\end{table*}

%---------- End Exp -------------------------

\if{0}
% \usepackage[normalem]{ulem}
\begin{table*}
\begin{minipage}{0.58\textwidth}
\scriptsize
\centering
\begin{tabular}{|l|l|l|l|l|l|l|}
\hline
\textbf{Model}  & \textbf{Architecture}             & \textbf{Acc}   & \textbf{M-F1}  & \textbf{F1 (H)} & \textbf{P (H)} & \textbf{R (H)} \\ \hline
\textbf{T1} & \textbf{fastText}              & 0.687          & 0.673          & 0.609           & 0.611          & 0.614          \\ \hline
\textbf{T2} & \textbf{LASER}                 & 0.730          & 0.720          & 0.668           & 0.655          & 0.686          \\ \hline
\textbf{T3} & \textbf{BERT}                  & 0.735          & 0.722          & 0.664           & 0.675          & 0.667          \\ \hline
\textbf{T4} & \textbf{HXP}                   & 0.757          & 0.733          & 0.653           & \textbf{0.753} & 0.577          \\ \hline
\textbf{A1} & \textbf{MFCC}                  & 0.675          & 0.665          & 0.622           & 0.593          & 0.679          \\ \hline
\textbf{A2} & \textbf{AVGG19}                & 0.690          & 0.669          & 0.589           & 0.629          & 0.559          \\ \hline
\textbf{V1} & \textbf{3D-CNN}                   & 0.674          & 0.653          & 0.571           & 0.619          & 0.547          \\ \hline
\textbf{V2} & \textbf{InceptionV3}                 & 0.720          & 0.706          & 0.643           & 0.653          & 0.637          \\ \hline
\textbf{V3} & \textbf{ViT}                   & 0.748          & 0.733          & 0.672           & 0.695          & 0.656          \\ \hline
\textbf{M1} & \textbf{BERT $\odot$ ViT $\odot$ MFCC} & \textbf{0.798} & \textbf{0.790} & \textbf{0.749}  & {\ul 0.742}    & \textbf{0.758} \\ \hline
\textbf{M2} & \textbf{BERT $\odot$ ViT $\odot$ AVGG19} & 0.755          & 0.765          & 0.718           & 0.723          & 0.719          \\ \hline
\textbf{M3} & \textbf{HXP $\odot$ ViT $\odot$ MFCC} & {\ul 0.777}    & {\ul 0.767}    & {\ul 0.720}     & 0.718          & {\ul 0.726}    \\ \hline
\textbf{M4} & \textbf{HXP$\odot$ ViT $\odot$ AVGG19} & 0.767          & 0.756          & 0.707           & 0.714          & 0.712          \\ \hline
\end{tabular}
    \caption{Model performance on the task of classification of hate videos. H: hate class, Acc: accuracy, M-F1: macro-F1, P: precision, R: recall, HXP: HateXplain.}
\label{tab:performanceMetric}
\end{minipage}
\begin{minipage}{0.35\textwidth}
\centering
\scriptsize
  \begin{tabular}{|l|c|c|}
\hline
\textbf{Model}               & \textbf{\begin{tabular}[c]{@{}c@{}}VL \textless{}= 105 sec\\  (542 videos)\end{tabular}} & \textbf{\begin{tabular}[c]{@{}c@{}}VL \textgreater 105 sec \\ (541 videos)\end{tabular}} \\ \hline
\textbf{T1} & 0.609                                                                                                       & 0.700                                                                                           \\ \hline
\textbf{T2} & 0.675                                                                                                       & 0.655                                                                                                                                \\ \hline
\textbf{T3} & 0.672                                                                                                       & 0.708                                                                                                                                \\ \hline
\textbf{T4} & 0.698                                                                                                       & 0.727                                                                                                                                \\ \hline
\textbf{A1} & 0.603                                                                                                       & 0.687                                                                                                                                \\ \hline
\textbf{A2} & 0.583                                                                                                       & 0.669                                                                                                                                \\ \hline
\textbf{V1} & 0.637                                                                                                       & 0.587                                                                                                                                \\ \hline
\textbf{V2} & 0.672                                                                                                       & 0.707                                                                                                                                \\ \hline
\textbf{V3} & 0.718                                                                                                       & 0.703                                                                                                                                \\ \hline
\textbf{M1} & \textbf{0.772}                                                                                              & \textbf{0.759}                                                                                                                       \\ \hline
\textbf{M2} & 0.743                                                                                                       & 0.733                                                                                                                                \\ \hline
\textbf{M3} & \underline{0.744}                                                                                                 & \underline{0.741}                                                                                                                          \\ \hline
\textbf{M4} & 0.733                                                                                                       & 0.731                                                                                                                                \\ \hline
\end{tabular}
\caption{Macro F1 Score of various models with respect to video length (VL) in secs.}
\label{tab:videoLenPerform}
\end{minipage}
\end{table*}

\fi

\section{Experiments and Results}
\label{sec:exptsRes}
\subsection{Experimental Setup}
\label{sec:experiments}

We evaluate our models using $k$-fold stratified cross-validation, which is beneficial in assessing models having less labeled data. For all the experiments, we set $k$ to 5 here, and for each fold, we use 70\% data for training, 10\% for validation, and the rest 20\% for testing. We use the same test sets across all the models to ensure a fair comparison. For all the unimodal neural network models, the internal layer has two fully connected layers of 128 nodes, reduced to a feature vector of length 2. For the uni-modal LSTM based models, all the frame embeddings are passed to an LSTM network with hidden size of 128, which is finally reduced to a feature vector of length 2.
For the 3D-CNN, each frame has been resized to 100 $\times$ 125. Both the Conv3D layers have 256 nodes, and the number of channels are 32 in the first layer, and 42 in the second layer. The kernel size for the first layer is $(5, 5, 5)$, and the second layer is $(3, 3, 3)$. Further, we use a stride of $(2, 2, 2)$ with zero padding for both the layers, resulting in a feature vector of length 2. We pass the final feature vector through a log-softmax layer with negative log-likelihood loss. This gives the probability of whether the video is hateful or not. For the fusion models, the text-based and audio based features are passed to two dense layers of size 128, which are finally fed to another dense layer of 64 nodes; the extracted vision-based features are passed to an LSTM network with a hidden size of 128, which is further passed to another dense layer of 64 nodes. Finally we concatenate all the nodes and reduce to a feature vector of length 2 as shown in Figure~\ref{fig:fusionModel}.
All the models are run for 20 epochs with Adam optimizer, batch\_size = 10, learning\_rate = $1e-4$. We store the results at the best validation score in terms of macro-F1 score. All models are coded in Python, using the Pytorch library.

\subsection{Evaluation Metric}
\label{sec:experiments}
To remain consistent with the existing literature, we evaluate our models in terms of the standard metrics -- accuracy, F1 score, precision, and recall. Together, these metrics should be able to thoroughly assess the classification performance of the models in distinguishing between the two classes -- hate vs non-hate. The best result is marked in \textbf{bold}, and the second best is \underline{underlined}.

\subsection{Results}\label{sec:results}

\subsubsection{Performance across different models}
Table~\ref{tab:performanceMetricNew}[Left side] shows the performance of each model.  We observe among all the text-based models, the transformer-based models perform the best, especially the \textbf{HateXPlain} model, which is earlier fine-tuned on a hate speech dataset. Among the audio-based models, we see \textbf{AudioVGG19} performs better than \textbf{MFCC}, though, in terms of macro-F1 score, the difference between these two models is marginal. For the vision-based models, we see the features extracted from \textbf{ViT} are very helpful in detecting hate videos among all other vision-based models. Further, we find that all the multi-modal models outperform all the unimodal models (in terms of accuracy, F1 score), and \textbf{BERT $\odot$ ViT $\odot$ MFCC} performs the best among all the models. %Please see Appendix~\ref{add:exp} for more results on an additional dataset.

\subsubsection{Performance based on video length}
We divide all the test datasets across all the folds in terms of video length. Empirically, we have the lower bucket with video length $\le105$ secs and the rest in the higher bucket to have almost the same number of videos across the two buckets. In Table~\ref{tab:performanceMetricNew}[Right Side] we report the macro F1 for both buckets. We observe that among the text-based models, except \textbf{T2}, all others perform better in the higher bucket. All the audio-based models, perform better in the higher bucket. On the other hand, except \textbf{V2}, all the vision-based models perform better in the lower bucket compared to their respective higher buckets. This is indicative that the vision-based model captures context well when the video duration is less.  When the text, audio, and vision-based models are integrated together, as expected the performance improved in both buckets.

\begin{figure}[h]
   \centering
    \includegraphics[width=0.7\linewidth]{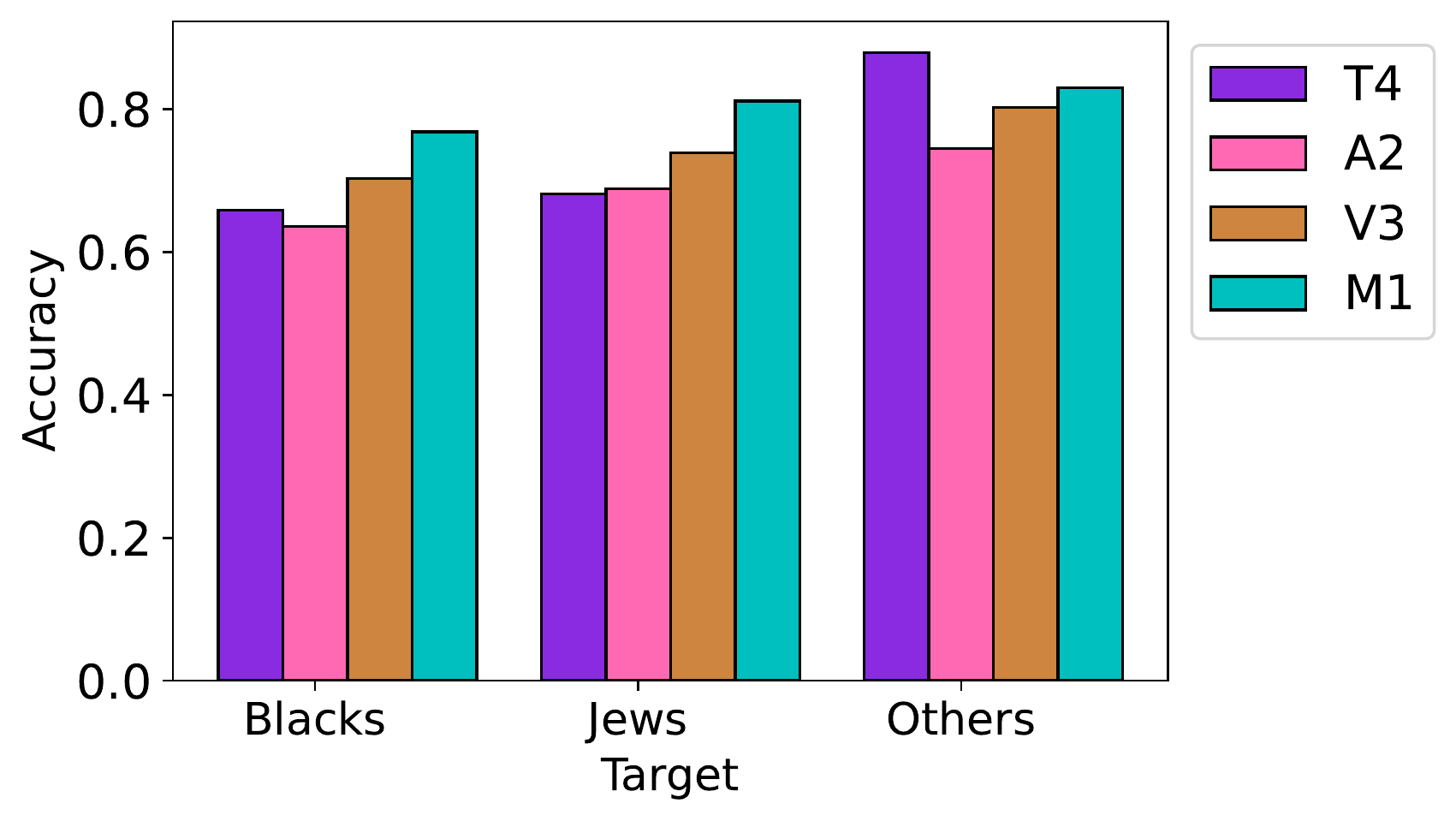}
    \caption{Target-wise performance. Only the best performing model for each modality and the best ensemble model are shown.}
    \label{fig:targetwisePerformace}
\end{figure}

\begin{figure}[h]
 \centering
    \includegraphics[width=0.7\linewidth]{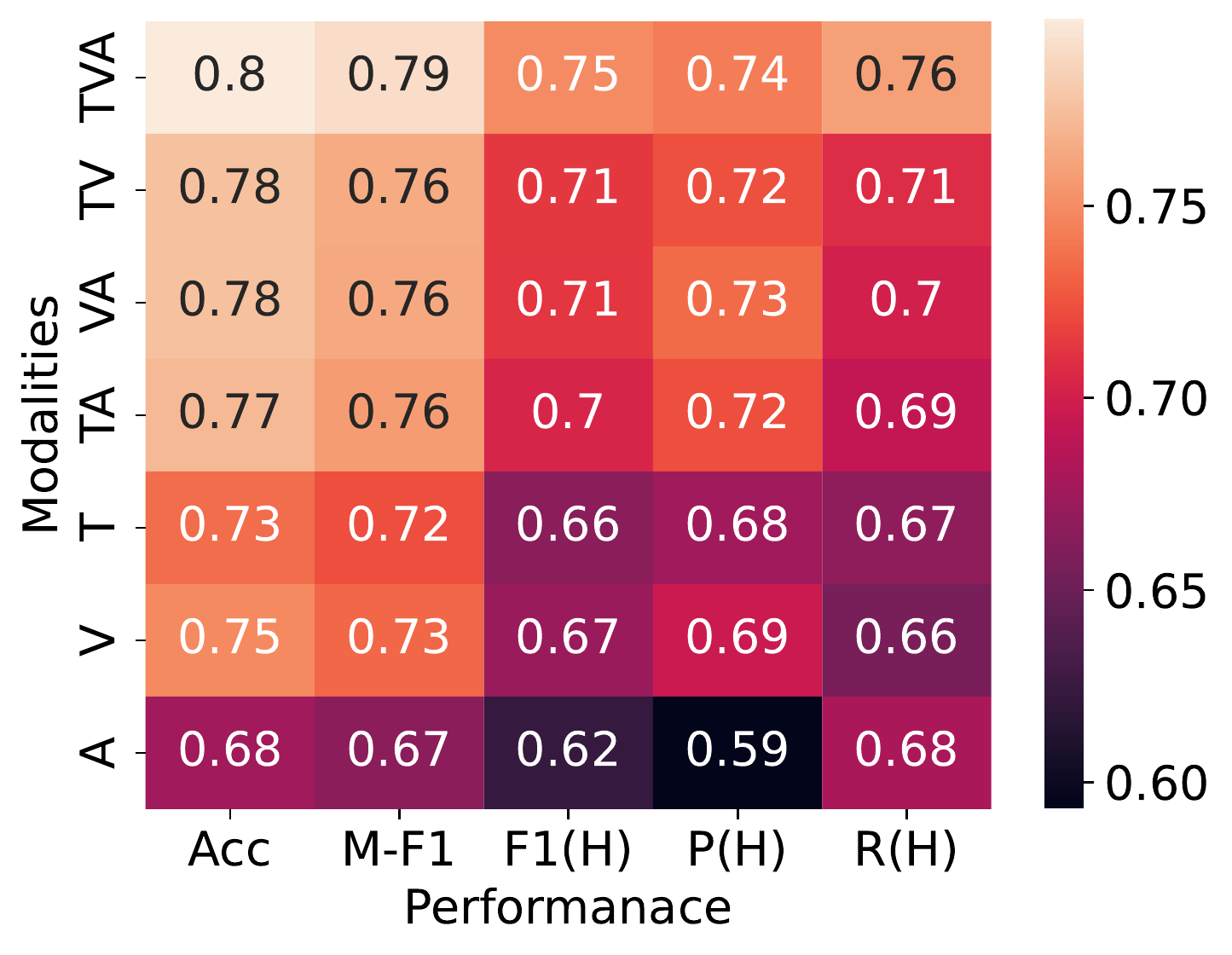}
    \caption{Heatmap of the performance of the different modalities. T: Textual, V: Vision, A: Audio.}
    \label{fig:heatmap}
\end{figure}

\begin{table*}[h]
\centering
\small
\begin{tabular}{|p{3.5cm}|p{4.2cm}|p{0.7cm}|p{6.8cm}|}
 \hline
\textbf{Video name} & \textbf{Description} & \textbf{Mode} & \textbf{Explanation} \\ \hline \hline
%\href{https://www.bitchute.com/video/FgsGCePHaxbq/}{
Terrorist Jew Hates Hollywood Traitor Kikes [ID=g11ysqwzlKj6] & In this video, a  person is seated and abusing Jews saying derogatory words like kikes.
 &   Text       & The video was unrelated to hate speech, and the transcript was clean; the audio-based model also failed due to the absence of high aggressiveness in the voice. \\ \hline
%\href{https://www.bitchute.com/video/lHk2E8MNU5HB/}{
Grinded Nig Freezer Full Of Ni**er Heads   [ID=lHk2E8MNU5HB] &  Here some group was yelling some type of song containing slur words like ``ni**er'' with nazi flag visible in the background           &    Audio + Video      &  The transcript was erroneous. Other modalities found useful signals based on nazi flag (video) and yelling (audio).    \\ \hline
%\href{https://www.bitchute.com/video/MuM1IdYxbDXe/}{
When Youre In Coon Town    [ID=OngXc0A4DXxo]          &  A song is yelled as a part of the audio about ``What is a c**n town?'' In the video, irrelevant images are shown.            &    Audio + Text      &  The audio models were able to capture it due to the presence of yelling. The transcript has derogatory words as signals. The video was fairly unrelated.    \\ \hline
%\href{https://www.bitchute.com/video/tpfjOeQu8tpS/}{
A Filthy Jew Straight From Hell Short Film  [ID=uSH9Z7tEj9vp]      &  In this video a person dressed in Nazi attire is abusing Jews people. &    Video + Text  & The video contained some hateful symbolism toward Jews and derogatory keywords were identified in the transcript, so both text and vision models succeeded.\\\hline
\end{tabular}
\caption{Examples of a few hate videos along with their description. We also mention the modality/ies which could predict the hate correctly in the \textit{Mode} column. In addition, we also provide a possible explanation for this prediction.}
\label{tab:manual_video}
\end{table*}

\subsubsection{Target-wise performance}
We also compute the target wise performance of the best uni-modal and fusion  models and show the results in  Figure~\ref{fig:targetwisePerformace}. We observe that among the uni-modal models the vision based model \textbf{V3} performs better for the videos targeting `Blacks' and `Jews', whereas the text-based model \textbf{T4} performs the best for the other targets (taken all together). Further, we notice that integrating these models, \textbf{M1}, gives consistently good performance across all the target communities.

\subsection{Effectiveness of the Modalities}
%\md{Need to discuss a bit how to rewrite, as there are a lot model combination}

To understand how the different modalities contribute to the prediction task, we perform ablation studies to demonstrate the effectiveness of all the modalities. We select our best multi-modal model \textbf{M1}, which utilizes all the features of a video for predicting the labels of the videos. We remove the modalities one at a time and train our models. We illustrate our result in Figure \ref{fig:heatmap} using a heatmap. We observe jointly training using all the modalities brings the highest performance. With the removal of at least one modality, performance drops by around 2-3\%. Further, with the removal of two modalities, we observe that the performance drops drastically. Overall, we find that the audio-based feature is successful when there was shouting or aggression in the voice present in the video because the MFCC features can capture these sound effects present in the audio. For example, one of the videos which showed a KKK\textsuperscript{\ref{footnoteKKK}} member shouting derogatory words is identified as hate by the audio but not by the other two modalities. The text-based model's performance depends on the accuracy of the automatic speech recognition (ASR). This model is successful most of the time when the ASR can correctly detect the hateful words in the transcript. Finally, the vision-based model is successful when it contains victims present in the video itself. There were some unrelated images in a few videos, like some game-play while the audio was derogatory. In such cases, the vision-based model fails due to the lack of useful signals. This also justifies the need for fusion models. We show a few examples of such videos in Table~\ref{tab:manual_video}.

\section{Conclusions and Future Work}
\label{sec:con}
This paper takes a step toward identifying hateful content in videos by leveraging signals across all three modes. To achieve this, we crawled videos from the BitChute platform and manually annotated them as hate and non-hate. Analyzing the annotated dataset \dataname{} revealed interesting aspects about the hate videos. We utilized all the modalities of the video to detect whether it is hateful or not. We showed that models which take multiple modalities into account performed better compared to the uni-modal variants. We also performed a preliminary analysis to understand how different modalities contribute to the prediction. We found that text-based model performs well when the transcript is clean, the audio-based model is successful when there is shouting or aggression in the video, and the vision-based model is able to capture the hateful content when hateful activities or the target of the abuse are present in the video.
 
In future we plan to use other vision and speech transformers such as ViViT~\cite{arnab2021vivit}, Wav2Vec~\cite{baevski2020wav2vec}, etc. which can possibly further boost the classification performance. One of the hardships here however is that we need much larger-sized videos to be annotated in order to train and fine-tune such data-hungry models.  Besides, we plan to annotate videos of longer length. We also envisage to build models which not only would detect a video as hateful but also identify the sections of the video which made it hateful. Thus, instead of looking into the full video, a moderator can watch the portions of the videos, which has been marked as hateful by the model and, subsequently, decide for moderation actions. %Such a pipeline will help in reducing the mental toll on the moderators as they will only need to see the relevant sections of the videos.

\section{Ethical Statement}
\subsection{Ethical Considerations}

Our database constitutes videos with labeled annotations and does not include any personally identifiable information about any user or the Bitchute channel where the videos have been uploaded. We only analyzed publicly available data. We followed standard ethical guidelines~\cite{rivers-ethical}, not making any attempts to track users across sites or deanonymize them. Since the video used in our analysis contain hateful elements, care should be taken to not use it for negative purposes like spreading further hatred or maligning an individual or a community.

\subsection{Biases}

Any biases noticed in the dataset are unintentional, and our intention is not bring any individual or a target community to harm. We believe it can be subjective to determine if a video is hateful or not; thus, biases in our gold-labeled data or label distribution are inevitable. Nonetheless, we are confident that the label given to the data is most accurate due to the significant inter-annotator agreement we have achieved.

\subsection{Intended Use}

We share our data to encourage more research on hate video classification. We only release the dataset for research purposes and do not grant a license for commercial or malicious use.

\bibliography{aaai23}

\end{document}